\documentclass[final,twocolumn,preprint]{elsarticle}

\usepackage{hyperref}
\usepackage{lineno,hyperref}
\modulolinenumbers[200]
\journal{Journal of Pattern Recognition}
\usepackage{epsfig}
\usepackage{graphicx}
\usepackage{amsmath}
\usepackage{amssymb}
\usepackage{subcaption}
\usepackage{multirow}
\usepackage{rotating}

\newcommand{\ie}{\textit{i.e.}}

\newcommand{\etal}{\textit{et al.~}}

\begin{document}
	
\begin{frontmatter}

\title{An Efficient Framework for Zero-Shot Sketch-Based Image Retrieval}

\author{Osman Tursun\corref{cor1}}
\cortext[cor1]{Corresponding author}
\ead{w.tuerxun@qut.edu.au}
\author{Simon Denman}
\ead{s.denman@qut.edu.au}
\author{Sridha Sridharan}
\ead{s.sridharan@qut.edu.au}
\author{Ethan Goan}
\ead{e.goan@hdr.qut.edu.au}
\author{Clinton Fookes}
\ead{c.fookes@qut.edu.au}
\address{Signal Processing, Artificial Intelligence and Vision Technologies (SAIVT) \\ Queensland University of Technology, Australia}

\begin{abstract}
Recently, Zero-shot Sketch-based Image Retrieval (ZS-SBIR) has attracted the attention of the computer vision community due to it's real-world applications, and the more realistic and challenging setting than found in SBIR. ZS-SBIR inherits the main challenges of multiple computer vision problems including content-based Image Retrieval (CBIR), zero-shot learning and domain adaptation. The majority of previous studies using deep neural networks have achieved improved results through either projecting sketch and images into a common low-dimensional space or transferring knowledge from seen to unseen classes. However, those approaches are trained with complex frameworks composed of multiple deep convolutional neural networks (CNNs) and are dependent on category-level word labels. This increases the requirements on training resources and datasets. In comparison, we propose a simple and efficient framework that does not require high computational training resources, and can be trained on datasets without semantic categorical labels. Furthermore, at training and inference stages our method only uses a single CNN. In this work, a pre-trained ImageNet CNN (\ie ResNet50) is fine-tuned with three proposed learning objects: \textit{domain-aware quadruplet loss}, \textit{semantic classification loss}, and \textit{semantic knowledge preservation loss}. The \textit{domain-aware quadruplet} and \textit{semantic classification} losses are introduced to learn discriminative, semantic and domain invariant features through considering ZS-SBIR as a object detection and verification problem. To preserve semantic knowledge learned with ImageNet and utilise it on unseen categories, the \textit{semantic knowledge preservation loss} is proposed. To reduce computational cost and increase the accuracy of the semantic knowledge distillation process, ground-truth semantic knowledge is prepared in a class-oriented fashion prior to training. Extensive experiments are conducted on three challenging ZS-SBIR datasets, Sketchy Extended, TU-Berlin Extended and QuickDraw Extended. The proposed method achieves state-of-the-art results, and outperforms the majority of related works by a large margin.

\end{abstract}


\end{frontmatter}

\linenumbers

\section{Introduction}

\begin{figure}[!th]
	\begin{center}
		\includegraphics[width=1\linewidth]{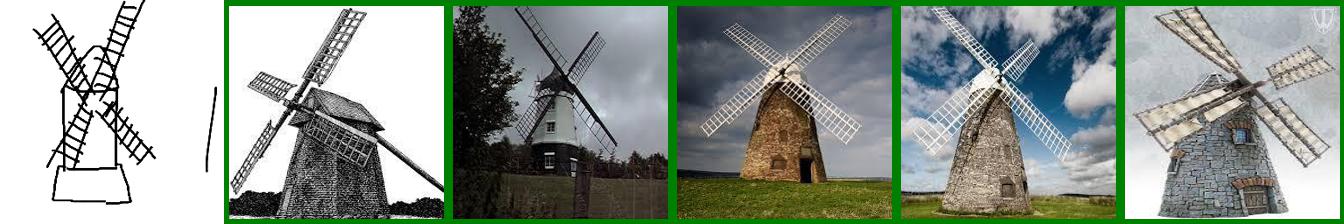}
		\includegraphics[width=1\linewidth]{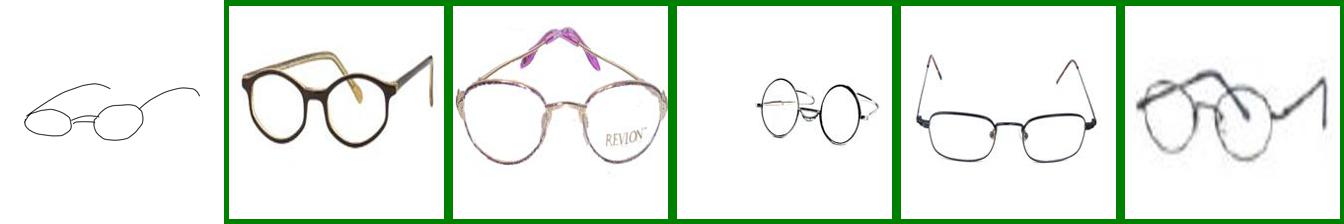}
		\includegraphics[width=1\linewidth]{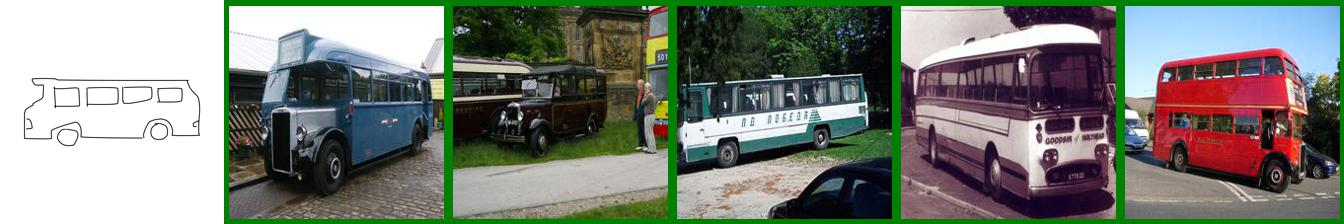}
		\caption{Examples of SBIR results. In each row, the figure in the left is query while others are retrieved results.}
		\label{fig:sbir}
	\end{center}
\end{figure}

Searching for images using an image query has increased in popularity as content-based image retrieval (CBIR) techniques have improved in recent years. However, the thrust of CBIR research has considered the scenario where both query and gallery images are real photos (\ie scenes \cite{gordo2016deep}, faces \cite{schroff2015facenet}) or digital images (\ie logo \cite{tursun2019component}). With the widespread popularity of touch-screen devices, free-hand sketch-based image retrieval (SBIR) tasks have drawn the attention of the computer vision (CV) community as sketches are a convenient, universal, easy and fast method for image description \cite{dey2019doodle,liu2019semantic,dutta2019style,dutta2020adaptive,wang2020deep,huang2018sketch}. See Figure \ref{fig:sbir} for examples of SBIR results.

The domain-gap and information-gap between sketch and photo domains presents a challenge to existing CBIR approaches. Sketches contain sparse and abstract information, while photos carry dense and precise information. Deep learning models have been applied to reduce these gaps, either in the latent space or the pixel space. A variety of complex architectures such as multiple independent networks \cite{zhang2016sketchnet,sangkloy2016sketchy,dey2019doodle,shen2018zero}, semi-hetergenous networks \cite{lei2019semi,liu2017deep}, generative adversarial networks (GAN) \cite{zhang2018generative,Verma_2019_CVPR_Workshops,pang2017cross,guo2017sketch} and networks with domain-invariant layers \cite{liu2019semantic,shen2018zero} have been proposed to address the domain gap. Although these approaches have shown substantial improvements over hand-crafted features, the increased model complexity requires extra resources during training and inference processes.

The model complexity has increased as SBIR methods are often evaluated under zero-shot settings, where testing queries are unseen during training. Related studies found existing SBIR models tend to fail under zero-shot settings \cite{shen2018zero}. To tackle this problem, commonly, the mapping or joint embedding space between the visual representation and class semantic representation is modelled. To achieve this, language models \cite{dey2019doodle,zhu2020ocean,Dutta2019SEMPCYC,zhang2020zero,Dutta2019SEMPCYC} are often used to extract semantic class embeddings, while auxiliary networks such as auto-encoders \cite{zhu2020ocean,Dutta2019SEMPCYC,dey2019doodle}, GANs \cite{Dutta2019SEMPCYC,Dutta2019SEMPCYC,li2019bi,pandey2020stacked} or graph convolutional neural networks (GNN) \cite{zhang2020zero,shen2018zero} are trained to learn the joined representation or mapping. 

Moreover, the high training cost is not the only drawback of the aforementioned methods. They require all classes of the training set have descriptive text labels that can be modelled by a language model. However, in some practical applications, classes may be only labelled with numerical values, or uncommon (\ie unknown) word labels which cannot be modelled by a language model.

This paper aims to tackle ZS-SBIR with a simple, efficient, and language model-free framework. The recent state-of-the-art (SOTA) work, SAKE \cite{liu2019semantic}, is a concise and simple framework. It's feature extraction encoder is a single-stream Convolutional Neural network (CNN) that ensures an efficient and simple inference process. However, during training it also requires a language model and another ImageNet pre-trained CNN to generate valid teacher signals for knowledge distillation. They argue preserving knowledge learned from Image-Net via knowledge distillation is beneficial for ZS-SBIR. Although we also find that rich features learned from Image-Net are essential for ZS-SBIR, we find a language model and online teacher network are not necessary. SAKE generates a teacher signal for each input item, either from the sketch or photo domain. It therefore requires a language model to align teacher signals that are otherwise invalid due to domain-shift. The alignment is based on the semantic similarity matrix of ImageNet labels and target dataset labels, which is constructed using WordNet \cite{miller1998wordnet}. In comparison, we generate teacher signals for each class by averaging the activations of pretrained ImageNet with images from the photo domain, whose distribution is close to ImageNet. We therefore do not require any semantic labels or language models, and the teacher signal generation is a one-time offline process.

We also find SAKE treats the ZS-SBIR problem as an object identification task, where the learning objective is a categorical classification loss, while other related works \cite{dey2019doodle,pandey2020stacked} consider the problem a verification task where metric learning \ie triplet loss, is applied. In this work, we not only unify these two objectives in a single framework, but also propose a domain-aware quadruplet loss for metric learning.

We have tested the proposed method on two popular SBIR datasets (Sketchy Extended \cite{sketchy2016} and TU-Berlin Extended \cite{eitz2012humans}) and a newly proposed challenging SBIR dataset, QuickDraw Extended \cite{dey2019doodle}. In all benchmarks, we have achieved state-of-the-art (SOTA) performance only through fine-tuning a ResNet50 \cite{he2016deep} model with our three proposed learning objectives: \textit{Domain-aware Quadruplet}, \textit{Semantic Classification} and \textit{Semantic Knowledge Preservation} losses.

The remainder of the paper is organized as follows. Section \ref{sec:lit} presents a literature review where we discuss related ZS-SBIR studies. In Section \ref{sec:met}, the proposed model and learning objectives are introduced. Section \ref{sec:exp} outlines experiment setups, results and related discussions; and finally Section \ref{sec:con} concludes the paper.

\section{Related Work}
\label{sec:lit}
Early SBIR studies mainly focus on the challenges raised by the large domain gap between the sketch and photo domain. Both hand-crafted features and deep features have been explored. Hand-crafted features include edge/shape-based features \cite{hu2013performance,saavedra2014sketch,saavedra2015sketch} with a bag-of-words representation, as in some aspects, strong edges in a photo correspond to the contours of sketches. On the other hand, deep features seek to learn a joint representation the of sketch and photo domain through metric learning \cite{yu2016sketch,qi2016sketch,song2017deep,liu2017deep}, style-content disentangle representation, \cite{dutta2019style,li2019bi} and style-transfer \cite{guo2017sketch,zhang2018generative,zhang2016sketchnet,bai2020cross}. However, related studies discover that the accuracy of these models decreases in a real-life and challenging scenarios where either the queries or gallery images are unseen. To tackle this problem, zero-shot SBIR approaches \cite{shen2018zero,liu2019semantic,dey2019doodle,zhu2020ocean,Dutta2019SEMPCYC,zhang2020zero,Dutta2019SEMPCYC} have been proposed.

The majority of ZS-SBIR approaches leverage semantic information embedded in seen data (\ie word labels) to learn a generalised representation for both seen and unseen categories. The main difference between these methods lies in the architecture of the mapping network and the embedding method in the semantic space. GNNs \cite{shen2018zero,zhang2020zero}, Multi Layer Perceptrons (MLP) \cite{dey2019doodle,chaudhuri2020simplified} and GANs \cite{deng2020progressive,Dutta2019SEMPCYC,xu2020correlated,zhu2020ocean} have all be used for the mapping network. On other side, word2vec \cite{dey2019doodle,zhu2020ocean,zhang2020zero,xu2020correlated,Dutta2019SEMPCYC,chaudhuri2020simplified,deng2020progressive} and hierarchical models \cite{zhu2020ocean,Dutta2019SEMPCYC,deng2020progressive} are common embedding methods used to construct the semantic space.

In comparison, the recent work SAKE \cite{liu2019semantic} uses a visual semantic representation learned from ImageNet. However, to avoid an incorrect representation caused by domain-shift, SAKE aligns the visual semantic representation with a semantic similarity matrix constructed with wordNet. We proposed an alternative method for extracting a visual semantic representation that is free-from alignment. As such, our method does not require a language model. To the best of our knowledge, making our proposed approach one of a very small number of ZS-SBIR studies that do not require a language model. Other such methods are typically generative approaches based-on GANs \cite{dutta2019style} and variational auto-encoder encoders (VAE) \cite{yelamarthi2018zero}.

\section{The Proposed Method}
\label{sec:met}
In this section, we present our proposed method for zero-shot sketch-based image retrieval (ZS-SBIR). In the following sub-sections, we first outline the overall structure of the proposed method, then explain network structures, learning objectives and discuss implementation details.

\begin{figure*}[!th]
	\centering
	\includegraphics[width=1\textwidth]{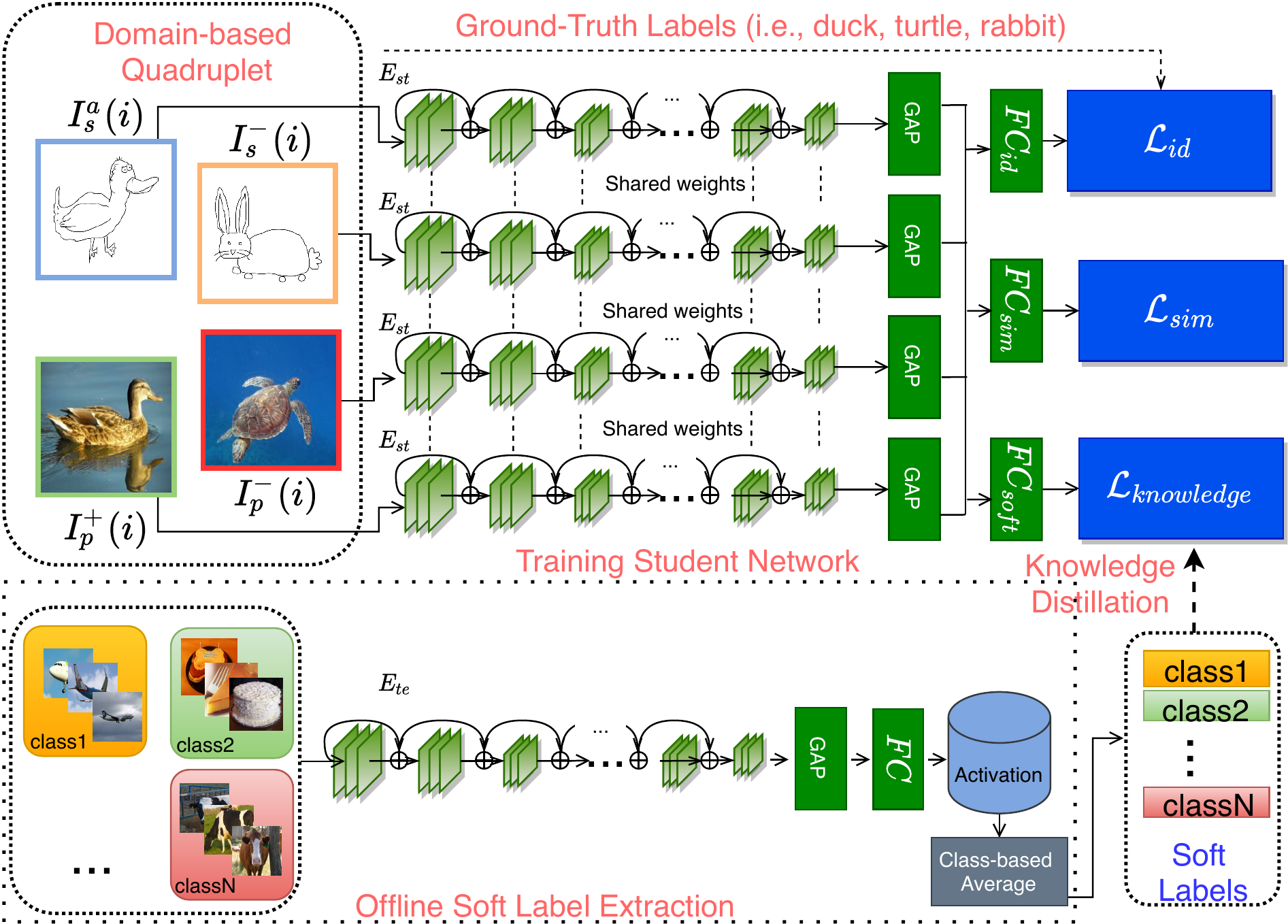}
	\caption{A diagram of the proposed method. The approach includes an \textit{Online Training Student Network} and an \textit{Offline Soft Label Extraction with Teacher Network}. The student, $E_{st}$, and teacher, $E_{te}$, networks use Resnet50 as the backbone. A quadruplet composed of two images from the sketch domain and two images from the photo domain is the input of the student network, while the inputs to the teacher network are only from the photo domain. The teacher network generates soft labels to prevent the student network from forgetting previously learned knowledge from ImageNet, and the soft label extraction is a one-time offline process.}
	\label{fig:diagram}
\end{figure*}

\subsection{Network Architecture}
The objective of the proposed method is to learn discriminative and domain-invariant CNN encoders that map semantically similar images from the sketch and photo domains into the same region of a common embedding space. An overall diagram of the proposed approach is shown in Figure \ref{fig:diagram}. The diagram is composed of two parts: the \textit{Online Training Student Network} and the \textit{Offline Soft Label Extraction with Teacher Network}. The student network, $E_{st}$, is trained with quadruplets and three proposed learning objectives. The teacher network, $E_{te}$, generates ground-truth for the knowledge distillation that prevents the $E_{st}$ from forgetting semantic knowledge learned from pre-training on ImageNet. Unlike \cite{liu2019semantic}, our approach does not require a teacher network during training and a language model for an alignment. Our approach, therefore, has a simple and efficient training process. For simplicity and efficiency, two encoders with ResNet50 \cite{he2016deep} backbones are used as $E_{st}$ and $E_{te}$. However, for $E_{st}$, we replace the fully-connected layer of ResNet50 with three new fully connected layers, $FC_{id}$, $FC_{sim}$ and $FC_{soft}$, that correspond to three proposed learning objectives. The size of $FC_{soft}$ is $1,000$, while sizes of $FC_{id}$ and $FC_{sim}$ are equal to the number of classes (\ie $80$, $100$, $104$ or $220$) and the size of embedded feature (\ie $64$, $512$ or $1024$). In this work, a four stream encoder where all streams share weights is used as $E_{st}$. However, semi-heterogeneous networks, or special domain-invariant layers widely used by previous works \cite{dey2019doodle, shen2018zero,liu2019semantic} to process the photos and sketches separately are easily integrated into $E_{st}$. Global average pooling (GAP) is applied to extract latent features from the last convolutional layer of the backbone network. Similar to \cite{dey2019doodle}, we have tried adding an attention \cite{xu2015show} mechanism to $E_{st}$, although we observed that it didn't yield any improvements during ablation studies.

\subsection{Learning Objectives}
To learn a discriminative and domain-invariant encoder with general semantic knowledge, we introduce the following learning objectives: \textit{Domain-aware Quadruplet Loss}, \textit{Classification Loss} and \textit{Knowledge Preservation loss}.\\
{\bfseries{Domain-aware Quadruplet Loss}} is a modified version of the triplet loss, which  has been widely used to maximise the inter-class distance and minimise the intra-class distance in embedding space for various image retrieval tasks \cite{gordo2016deep,chen2017beyond}. Here, our objective is also minimizing distance between sketches and photos from the same semantic category, while maximizing the distance between sketches and photos from different categories in the target embedding space. 

With the triplet loss, this inter-class and intra-class distance relationship is formulated with triplets where a sketch and a photo are selected from the same category, while another photo is from a different category. For example, $T(i)=\left\lbrace I^a_s(i), I_p^+(i), I_p^-(i)\right\rbrace$ is $i$th triplet where $l(I_s^a(i)) = l(I_p^+(i))$ and $l(I_s^a(i)) \neq l(I_p^-(i))$ (notation $l$ represents label). The Euclidean distance between the anchor sketch and the positive (same class) photo image is $\delta^+(i) = ||E_{st}(I_{s}^a(i)) - E_{st}(I^+_{p}(i))||_2^2$, while the Euclidean distance between the anchor sketch and it's negative (different class) photo image is $\delta^-_p(i) = ||E_{st}(I_s^a(i)) - E_{st}(I_p^-(i))||_2^2$. $\delta^-_p(i)$ should be larger than $\delta^+(i)$ by a threshold $\alpha$, which is set to 0.2. The triplet loss for a batch of $N$ triplets is defined as 

\begin{align}
\mathcal{L}_{sim}= \frac{1}{N} \sum_{i=1}^{N} max(\delta^+(i) - \delta_p^-(i) + \alpha, 0). %
\end{align}

The proposed domain-aware quadruplet loss deploys an extra negative sketch image, $I_s^-$, such that the quadruplet is defined as $Q=\left\lbrace I_s^a, I_p^+, I_p^-, I_s^- \right\rbrace$. The additional image is used to calculate the Euclidean distance between the anchor sketch and the additional negative (different class) sketch image, $\delta^-_s(i) = ||E(I_s^a(i)) - E(I_p^-(i))||_2^2$. Therefore, the proposed loss is

\begin{align}
	\mathcal{L}_{sim} = \frac{1}{2N} \sum_{i=1}^{N}(max(\delta^+(i) - \delta_p^-(i) + \alpha, 0) + \\ max(\delta^+(i) - \delta_s^-(i) + \alpha, 0)).
\end{align}

We proposed the quadruplet for the following reasons: 
\begin{enumerate}
    \item To overcome domain imbalance which can appear in triplet loss and classification losses (discussed later), as the total number of sampled photos are two times the number of sketches.
    \item Related studies \cite{chen2017beyond,khatun2020joint} demonstrate that an extra negative image is beneficial for learning discriminative features. However, these studies do not take consider domain differences and imbalance in their formulations.
\end{enumerate}

\noindent{\bfseries{Semantic Classification Loss}} is introduced to ensure hidden features extracted with $E_{st}$ are composed of signals that are sufficient to identifying the semantic classes of inputs from both the sketch and photo domains. Additionally, with this semantic loss, $E_{st}$ implicitly learns to minimise the intra-class distance. Specifically, a soft-max cross-entropy loss is utilised. As Equation \ref{eq:id} shows, every input to $E_{st}$ is a quadruplet, $Q$, that includes two images from the sketch domain and two images from the photo domain. This equal domain sampling ensures domain balance. The output from $E_{st}$ will be sent to the $FC_{cls}$ for softmax calculation. Here, we simply use notation $\phi$ to represent this whole process,
\begin{equation}
\label{eq:id}
\mathcal{L}_{cls} = -\frac{1}{4N}\sum_{i=1}^{N}\sum_{I\in Q}-logp(l(I(i))|\phi(I(i))), %
\end{equation}
where $p$ represents the probability.

\noindent{\bfseries{Semantic Knowledge Preservation Loss}} Transfer learning plays a key role in SBIR tasks. Networks pre-trained on ImageNet have been fine-tuned for ZS-SBIR problems in previous works \cite{shen2018zero,dutta2019style,dey2019doodle,liu2019semantic}. However, Liu \etal \cite{liu2019semantic} claim fine-tuning will cause catastrophic forgetting that decreases the ability of the fine-tuned network to adapt back to the original domain. To prevent a network from forgetting previously learned knowledge, Liu \etal generates a teacher signal to each of the training inputs for knowledge distillation. However, this requires extra training resources as inputs are also sent to a teacher network to generate the teacher signals. Moreover, their method also requires a language model for error alignments. Here, we implement a similar knowledge distillation approach, which is efficient and does not require a language model. As shown in Figure \ref{fig:diagram}, we only use a class-based teacher signal rather than item-based teacher signals. The teacher signals are the softmax of the average activation of the teacher network $E_{teacher}$ for each semantic class. The teacher signals can be considered as soft signals. The notation $q(l(I))$ represents the soft label of image $I$. To reduce the errors caused by domain shifts, we calculated $q$ with the softmax of the average activation of each class that exists in the photo domain as shown in Figure \ref{fig:diagram}. Those soft labels are only calculated once, so it is efficient. We use the cross-entropy loss with soft labels for calculation of the Knowledge loss $\mathcal{L}_{knowledge}$

\begin{equation}
	\mathcal{L}_{knowledge} = -\frac{1}{4N}\sum_{i=1}^{N}\sum_{I\in Q}-q(l(I(i))log\sigma(E(I(i))
\end{equation}

In summary, $E_{st}$ is trained using the $\mathcal{L}$ in Equation \ref{eq:fl}, which is a combination of the three proposed objectives. For simplicity, the weights of each objective are set to 1.

\begin{equation}
	\mathcal{L} = \mathcal{L}_{knowledge} + \mathcal{L}_{cls} + \mathcal{L}_{sim}
	\label{eq:fl}
\end{equation}

\subsection{Implementation Details}
PyTorch \cite{NEURIPS2019_9015} is used as our implementation framework, and all models are trained with single GTX 1080Ti GPU. We select an ImageNet pretrained ResNet50 as the backbone for both teacher and student networks. We applied the SGD optimiser with momentum=0.9 and decay=$5\times10^{-4}$. The batchsize is 16, but it includes 64 images as each input is a quadruplet. The initial learning rate $\lambda=1\times10^{-4}$, and it is decayed by a factor of 10 times after every ten epochs. We trained all models for up to 25 epochs, which is smaller than what previous works \cite{dey2019doodle,li2019bi,liu2019semantic} require, as our model starts to converges after only a few training epochs. We also used early-stop based on the validation accuracy. If the model's validation accuracy has not shown improvements within 5 epochs, the model will stop training.
\section{Experiments}
\label{sec:exp}
\subsection{Datasets}
We evaluated our method on well-known large-scale SBIR datasets: \textit{Sketchy Extended}, \textit{TU-Berlin Extended} and \textit{QuickDraw Extended}. An overall comparison of these datasets is described in Table \ref{tab:dataset-comp}.

{\noindent \bfseries{Sketchy Extended}} is an extended version of the Sketchy dataset \cite{sketchy2016} by Liu \etal \cite{liu2017deep}. The Sketchy dataset has $125$ categories. Each category is composed of 100 natural images and at least $600$ sketches. It's photo domain is extended by adding an extra $60,502$ natural images collected from ImageNet. The extended version has an average of $604$ sketches and $584$ images in each class, and it is a balanced dataset as the variance between the number of items in each class is relatively small. To adapt this dataset for zero-shot studies, the dataset is partitioned into seen and unseen sets. There exists two partition protocols in the literature. For clarify, we refer to them as \textit{SK-SH} and \textit{SK-YE}. SK-SH is proposed by Shen \etal \cite{shen2018zero}, who creates an unseen set by randomly selecting $25$ classes, and the remaining $100$ classes are used as training classes. However, some of those randomly selected classes might have already been seen by networks initialised with ImageNet pretrained weights, and thus this violates the zero-shot setting. SK-YE introduced by Yelamarthi \etal \cite{yelamarthi2018zero}. They carefully selects $21$ classes that are not present in ImagenNet.

{\noindent \bfseries{TU-Berlin Extended}} includes $20,000$ sketches from the TU-Berlin dataset \cite{eitz2012humans} and an extra $204,489$ real images collected by Liu \etal \cite{liu2017deep}. Its sketch-domain has a uniform class distribution but with only $80$ items, while the photo-domian has around $787$ items, but is highly imbalanced. It, therefore, is a challenging dataset. The partition protocol introduce by Shen \etal \cite{shen2018zero} is used for creating zero-shot training and testing sets. We refer to this protocol as \textit{TUB-SH} where $30$ randomly picked classes that include at least $400$ photo images are used for testing, and other classes are used for training.

{\noindent \bfseries{QuickDraw Extended}} is a challenging dataset created by Dey \etal \cite{dey2019doodle}. Compared to Sketchy Extended and TU-Berlin Extended datasets, it includes more sketches (average of $3022$/class) and photos (average of $1853$/class). All sketches are drawn by amateurs, so they are very abstract and highly variable. Moreover, all classes are carefully selected to avoid ambiguity and overlap. A partition following a similar protocol to that proposed by Yelamarthi \etal \cite{yelamarthi2018zero} is provided. We named this partition \textit{QD-DE}. With this partition, the dataset is split into $80$ training and $30$ testing classes.

\begin{table*}[!t]
    \caption{Comparison of public SBIR datasets. These datasets include images from the sketch and photo domains. For zero-shot studies, they are split to train (seen) and test (unseen) classes.}
    \label{tab:dataset-comp}
	\centering
	\begin{tabular}{c | c |c | p{2cm}| p{2cm}} 
		\hline
		 & \multicolumn{2}{c|}{\bf Sketchy Ext. \cite{liu2017deep}}    & \bf TU-Berlin Ext. \cite{zhang2016sketchnet}  & \bf QuickDraw Ext. \cite{dey2019doodle}           \\ \hline
		\bf \# Sketch/Class & \multicolumn{2}{c|}{604 $\pm$ 61} & 80 $\pm$ 0    & 3022 $\pm$ 216         \\ \hline
		\bf \# Image/Class  & \multicolumn{2}{c|}{584 $\pm$ 76} & 787 $\pm$ 489 & 1853 $\pm$ 308         \\ \hline
		\bf Name         & SK-SH \cite{shen2018zero} & SK-YE \cite{yelamarthi2018zero}                                      & TUB-SH \cite{shen2018zero}  & QD-DE \cite{dey2019doodle} \\\hline
		\bf \multirow{2}{*}{Type}         & 		\bf \multirow{2}{*}{Random} & 		\bf \multirow{2}{*}{ImageNet}                         & 		\bf \multirow{2}{*}{Random}      & 		\bf \multirow{2}{*}{ImageNet} \\
		         & 		 & 		\bf orthogonal                         & 	      & 		\bf orthogonal \\\hline
		\bf \# Train Class        & 100    & 104                                         & 220         & 80 \\\hline
		\bf \# Test Class        & 25     & 21                                          & 30          & 30 \\\hline
	\end{tabular}
\end{table*}

\subsection{Evaluation metrics}
Precision (P) and mean average precision (mAP) are two main metrics for evaluating the ranked retrieval results for testing queries in related SBIR studies. Precision is calculated for the top $k$ (\ie, 100, 200) ranked results, and mAP values are calculated for the top $K$ or all ranked results. The P@K is equal to the ratio between the number of total documents and relevant documents in the $K$ retrieved results. P@K is also used for calculating AP values of each query as follows:

\begin{equation}
	AP@K = \sum_{i=1}^{K}\frac{P@i \times \gamma(i)}{N},
\end{equation}

where $N$ is total number of relevant documents and $\gamma(i)$ is $1$ if the $i$th ranked result is relevant, otherwise $0$. mAP@k is mean AP@k of all queries.

\begin{table*}[!h]
	\centering
	\caption{A performance comparison of recent state-of-the-art ZS-SBIR methods.}
	\begin{tabular}{l | c | c c | c c | c c} 
		\hline
		\multirow{2}{*}{\bf Method} & \multirow{2}{*}{\bf DIM.} & \multicolumn{2}{p{2cm}|}{\bf Sketchy Ext. (Split: SK-SH)} & \multicolumn{2}{p{2cm}|}{\bf Sketchy Ext. (Split: SK-YE)} & \multicolumn{2}{p{2cm}}{\bf TUBerlin Ext. (Split: TU-SH)} \\\cline{3-8}
		&  & mAP & P  & mAP & P & mAP & P\\
		&  & @all & @100  & @200 & @200 & @all & @100\\\hline
		
		ZSIH \cite{shen2018zero} & $64^*$ & 25.8 & 34.2 & - & - & 22.3 &  29.4   \\ %
		\multirow{2}{*}{EMS \cite{lu2018learning}} & $64^*$ & - & - & - & - & 16.5 &  25.2 \\
		                                      & 512 & - & - & - & - & 25.9 &  36.9   \\
		CVAE \cite{yelamarthi2018zero} & 4,096 & 19.6 & 28.4 & 22.5 & 33.3 & - &  -\\
		GZS-SBIR \cite{Verma_2019_CVPR_Workshops} & 2,048 & 28.9 & 35.8 & - & - & 23.8 &  33.4\\
		\multirow{2}{*}{SEM-PCYC \cite{Dutta2019SEMPCYC}} & 64 & 34.9 & 46.3 & - & - & 29.7 &  42.6\\
		                                 & $64^*$ & 34.4 & 39.9 & - & - & 29.3 & 39.2\\
		Doodle2Search \cite{dey2019doodle} & 4,096 & - & - & $46.1^a$ & 37.0 & 10.9 &  - \\
		SketchGCN \cite{zhang2020zero} & 2,048 & - & - & $56.8^a$ & 48.7 & 32.4 &  50.5  \\
		Style-guide \cite{dutta2019style} & 4,096 & 37.6 & 48.4 & 35.8 & 40.0 & 25.4 &  35.6\\
		\multirow{2}{*}{SAKE \cite{liu2019semantic}} & 64* & 36.4 & 48.7 & 35.6 & 47.7 & 35.9 & 48.1\\
		                                             & 512 & 54.7 & 69.2 & 49.7 & 59.8 & 47.5 & 59.9 \\
		BDT \cite{li2019bi} & 1024 & - & - & 28.1 & 39.7 & - &  - \\
		OCEAN \cite{zhu2020ocean} & 512 & 46.2 & 59.0 & - & - & 33.3 &  46.7 \\
		\multirow{2}{*}{PCMSN \cite{deng2020progressive}} & 64 &  52.3 &  61.6  & -  & -  & 42.4 & 51.7    \\
		                & $64^*$ &  50.6 &  61.5  & -  & -  & 35.5 & 45.2    \\\hline
		SBTKNet & 512 & \bf 55.25 & \bf 69.77 & \bf 50.2 & \bf 59.6 & \bf 48.0 & \bf 60.8 \\\hline
	\end{tabular}
	\label{tab:rst-sktu}
	\\
	\footnotesize{$^a$ These mAP@200 evaluations use a different formulation to ours. If we follow the same mAP@200 evaluation protocol, our mAP@200 values for Sketchy-Ext. (Split: SK-YE) is 72.24.}\\
\end{table*}

\begin{table*}[!ht]
	\centering
	\caption{Comparison results of generalised ZS-SBIR on Sketchy Extended and TU-Berlin Extended datasets.}
	\begin{tabular}{c | c | c c | c c } 
		\hline
		\multirow{2}{*}{\bf Method} & \multirow{2}{*}{\bf DIM.} & \multicolumn{2}{p{2cm}|}{\bf Sketchy Ext. (Split: SK-SH)} & \multicolumn{2}{p{2cm}}{\bf TU-Berlin Ext. (Split:TU-SH)} \\\cline{3-6}
		&  & mAP@all & P@100  & mAP@all & P@100\\\hline
		ZSIH \cite{shen2018zero} &  64 &  21.9 &  29.6  &  14.2 & 21.8\\
		SEM-PCYC \cite{Dutta2019SEMPCYC} &  64 & 30.7 &  36.4  & 19.2 & 29.8 \\
		Style-guide \cite{dutta2019style} & 4,096 & 33.1 & 38.1 & 14.9 & 22.6 \\
		Ours & 512 & \bf 51.45  & \bf 57.20 & \bf 33.4   & \bf 49.4\\\hline
	\end{tabular}
	\label{tab:rst-gzs}
\end{table*}

\begin{table}[!ht]
	\centering
	\caption{Comparison results of QuickDraw-Extended Dataset.}
	\begin{tabular}{c | c | c c} 
		\hline
		\bf Method & \bf DIM. & \bf  mAP@all & \bf P@200 \\\hline
		CVAE \cite{yelamarthi2018zero} & 4,096 & 0.30 & 0.30 \\
		Doodle2Search \cite{dey2019doodle} & 4,096 & 7.52 & 6.75  \\\hline
		Ours & \bf 512 & \bf 11.88 & \bf 16.65\\\hline
	\end{tabular}
	\label{tab:rst-qd}
\end{table}

\subsection{State-of-the-art Comparison}
We have compared the proposed methods with SOTA methods on ZS-SBIR and its generalised version (search space includes \textit{seen} and \textit{unseen} categories \cite{Dutta2019SEMPCYC}) . The results of the ZS-SBIR task on the Sketchy Extended and TU-berlin Extended datasets are shown in Table \ref{tab:rst-sktu}, and the ZS-SBIR results on the QuickDraw Extended datasets are listed in Table \ref{tab:rst-qd}. The results of the generalised ZS-SBIR task on the Sketchy Extended and TU-Berlin Extended datasets are shown in Table \ref{tab:rst-gzs}.

In all these experiments, we have shown improvements compared to SOTA methods. We have surpassed methods that have not utilized a language model by a large margin. We also compare the embedding feature sizes used by the methods. Our feature size is $512$, which we note is relatively small compared to many other methods, and equal to feature size of the previous best SOTA method, SAKE \cite{liu2019semantic}. 

\subsection{Qualitative Results}
\begin{figure*}[!htbp]
\noindent\begin{subfigure}{.45\textwidth}
\centering
\includegraphics[width=\linewidth]{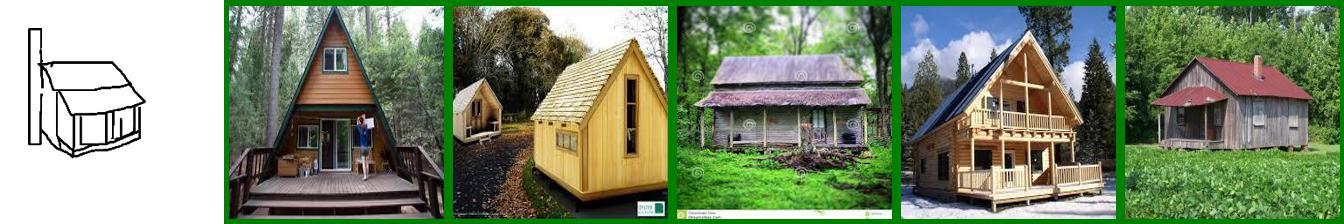}
\includegraphics[width=\linewidth]{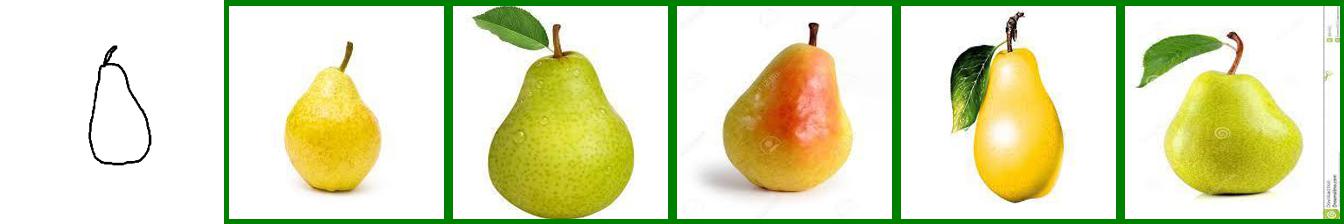}
\includegraphics[width=\linewidth]{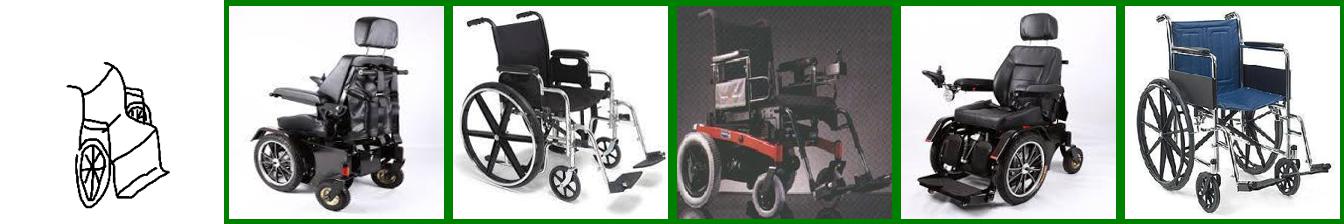}
\includegraphics[width=\linewidth]{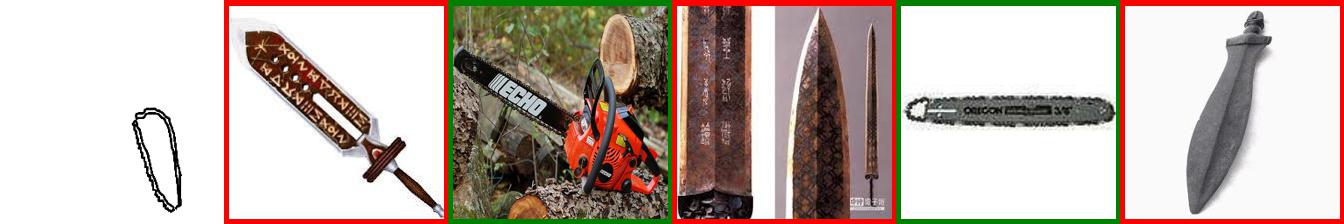}
\captionof{figure}{Zeroshot, Sketchy (Split:SK-YE)}
\label{fig:qual1}            
\end{subfigure}%
\hfill
\begin{subfigure}{.45\textwidth}
\centering
\includegraphics[width=\linewidth]{images/qualitative_results/zeroshot/sketchy/1000}
\includegraphics[width=\linewidth]{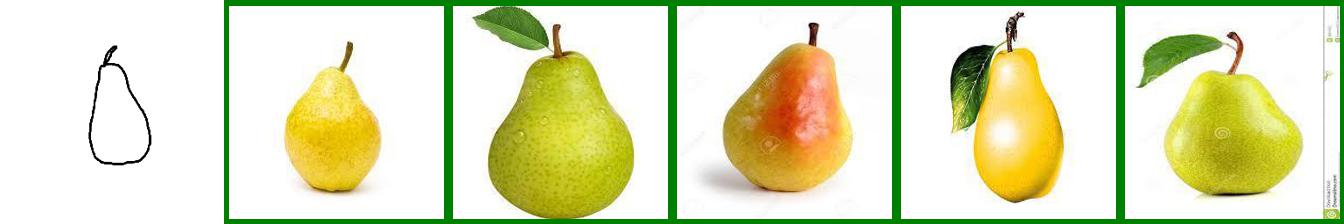}
\includegraphics[width=\linewidth]{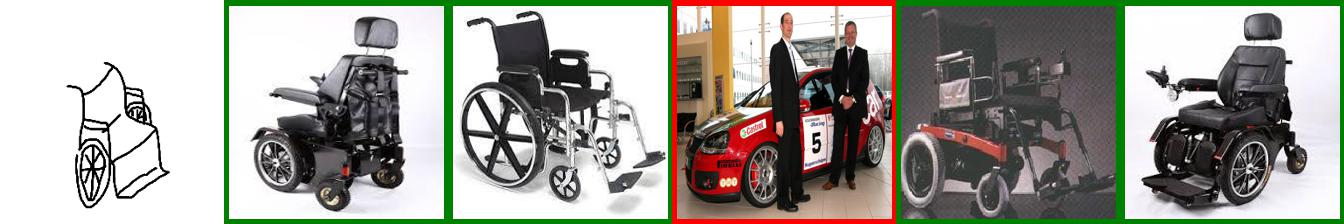}
\includegraphics[width=\linewidth]{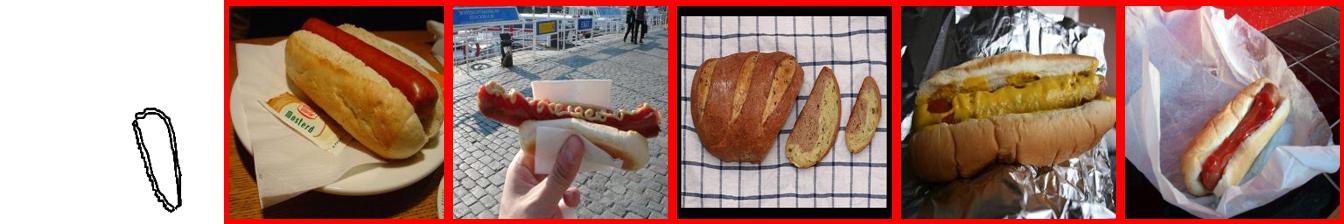}
\captionof{figure}{Generalized zeroshot, Sketchy (Split:SK-YE)}
\label{fig:qual2}            
\end{subfigure}%

\noindent\begin{subfigure}{.45\textwidth}
	\centering
	\includegraphics[width=\linewidth]{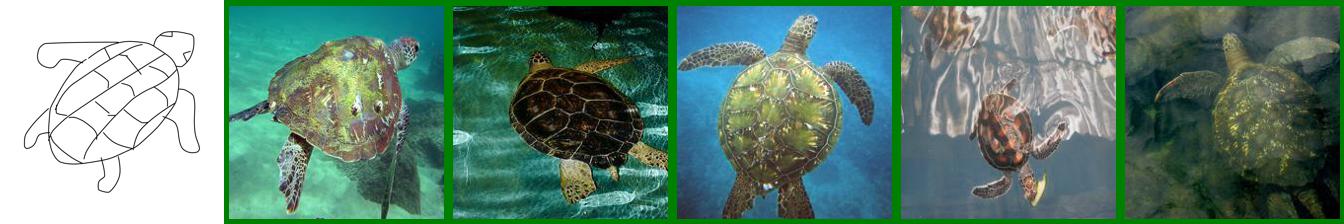}
	\includegraphics[width=\linewidth]{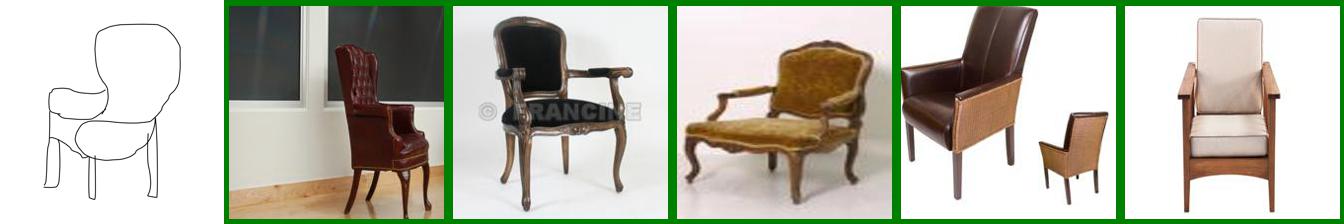}
	\includegraphics[width=\linewidth]{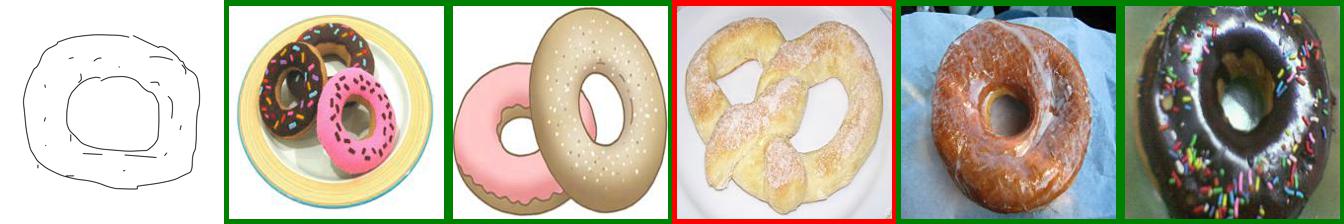}
	\includegraphics[width=\linewidth]{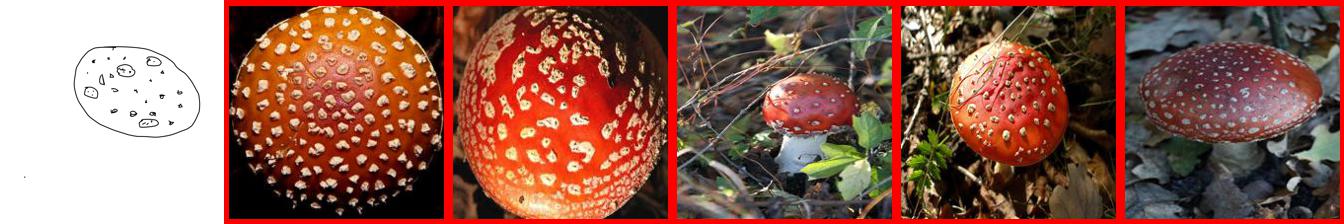}
	\captionof{figure}{Zeroshot, TU-Berlin (Split:TUB-SH)}
	\label{fig:qual3}            
\end{subfigure}%
\hfill
\begin{subfigure}{.45\textwidth}
\centering
\includegraphics[width=\linewidth]{images/qualitative_results/zeroshot/tuberlin/900}
\includegraphics[width=\linewidth]{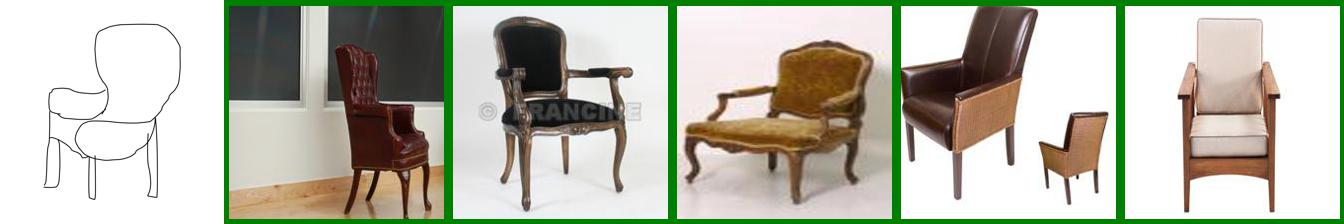}
\includegraphics[width=\linewidth]{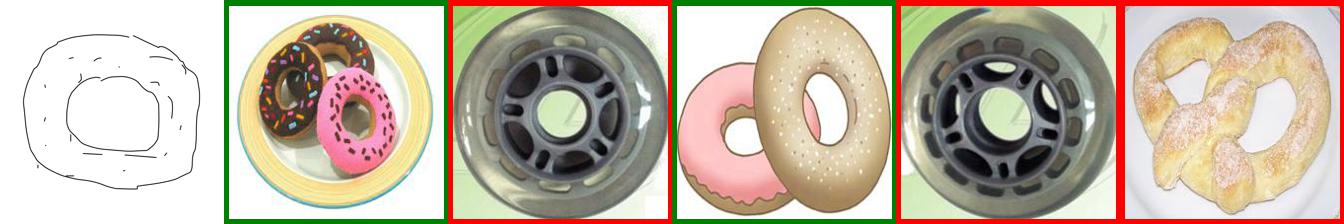}
\includegraphics[width=\linewidth]{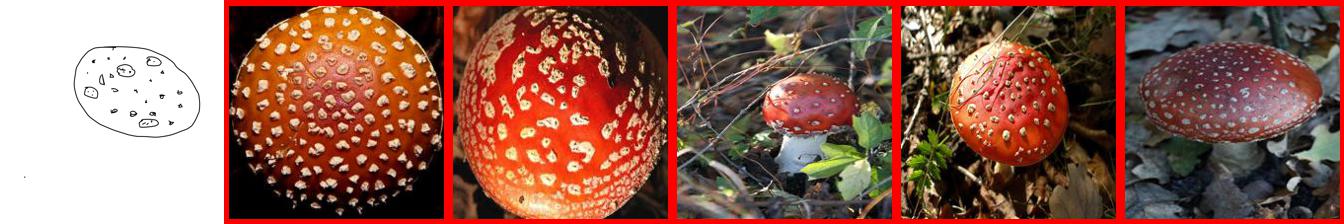}
\captionof{figure}{Generalized zeroshot, TU-Berlin (Split:TUB-SH)}
\label{fig:qual4}            
\end{subfigure}%
\caption{Top-5 ZS-SBIR (a,c) and  generalised ZS-SBIR (b,d) results retrieved by our model on Sketchy Ext. (a,b) and TU-Berlin Ext. (c,d) datasets. Correct results are shown with a green border, while false results are shown with a red border. The top two rows are all correct, the third row is partially correct, while the bottom row is all incorrect.}
\label{fig:qual-sktu}
\end{figure*}

\begin{figure*}
\centering
\includegraphics[width=1\linewidth]{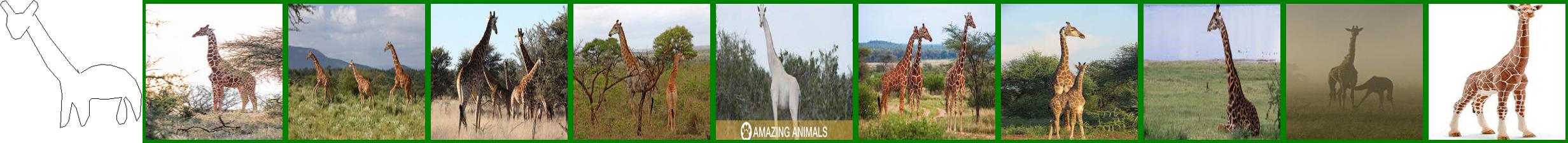}
\includegraphics[width=1\linewidth]{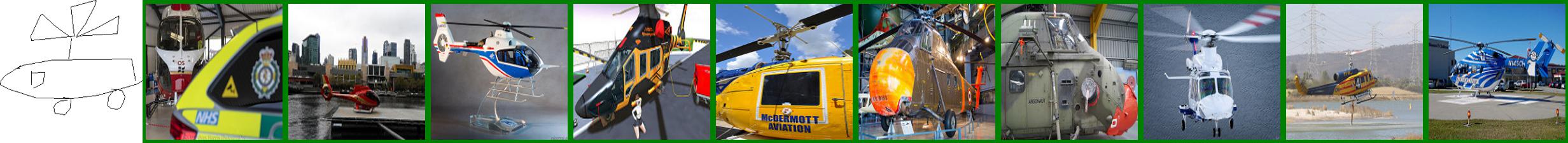}
\includegraphics[width=1\linewidth]{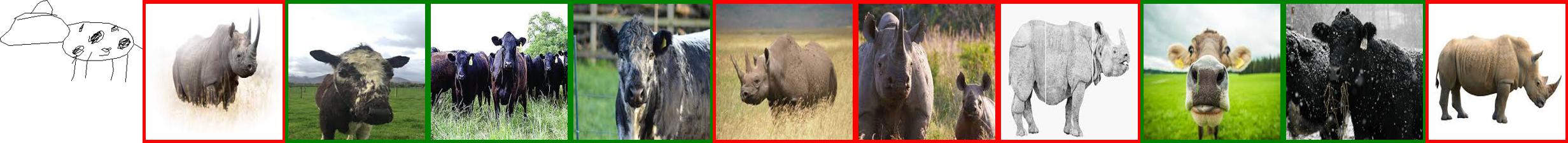}
\includegraphics[width=1\linewidth]{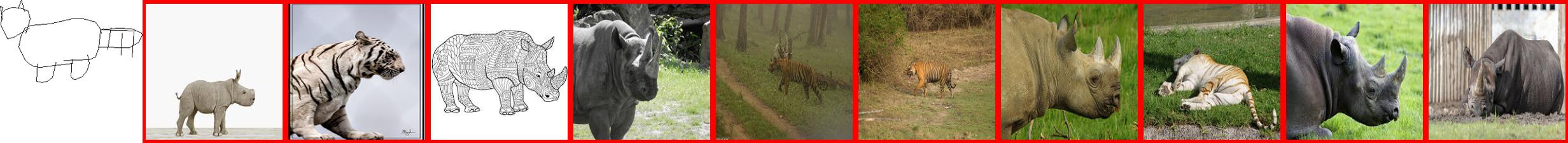}
\caption{Top-10 ZS-SBIR results retrieved by the proposed model on the QuickDraw Ext. dataset. Correct results are shown with a green border, while incorrect results are shown with a red border. The top two rows are all correct, the third row is partially correct, while the bottom row is all incorrect.}
\label{fig:qual-qd}            
\end{figure*}

We also visualized top10 and top5 results of some success and failure cases. All those results are displayed in Figure \ref{fig:qual-sktu} and Figure \ref{fig:qual-qd}. Figure \ref{fig:qual-sktu} displays both ZS-SBIR and GZS-SBIR results of Sketchy Ext. and TU-Berlin Ext. datasets. While Figure \ref{fig:qual-qd} shows top10 results of QuickDraw Ext. dataset. The proposed method returns perfect results when the given query is unambiguous, whereas it returns acceptable false-positive results when the query is unclear. This is more like to happen when the searching space is as large as in generalised ZS-SBIR cases.

\subsection{Ablation Studies}
Here, we investigate the impact of each proposed learning objective and attention (att.) \cite{xu2015show} on the proposed approach using the TU-Berlin Extended and Sketchy Extended Datasets. As shown in Table \ref{tab:ablation}, we provided results of models trained with several combinations of these losses. The model trained with the triplet loss is the baseline, and it's results show that it is a challenging baseline which outperforms most of the state-of-the-art method listed in Table \ref{tab:rst-sktu}. Each learning objective improves the results of the baseline. However, the attention mechanism has not brought any improvements to the final results. We supposed that, with the proposed learning objectives, the network is trained to pay attention to important information without an attention module.

\begin{table*}[!ht]
	\small
	\caption{Ablation study for the proposed approach. The same backbone network trained with the triplet loss is used as baseline.}
	\centering
	\label{tab:ablation}
	\begin{tabular}{c c c c|c c c c|c c c c}
		\hline
		\bf \multirow{4}{*}{Quad.} & \bf \multirow{4}{*}{ID.} & \multirow{4}{*}{\bf Know.} & \multirow{4}{*}{\bf Att.} & \multicolumn{4}{p{3cm}|}{\bf Sketchy Ext. (Split: SK-YE)} & \multicolumn{4}{p{3cm}}{\bf TUBerlin Ext. (Split: TU-SH)} \\\cline{5-12}
		& & & & mAP & mAP & P & P & mAP & mAP & P & P \\
		& & & & @all & @200 & @100 & @200 & @all & @200 & @100 & @200 \\\hline
		- & - & - & - & 47.7 & 43.6 & 55.1 & 52.0 & 44.8 & 46.2 & 56.8 & 55.0 \\
		\checkmark & - & - & - & 48.5 & 44.4 & 56.0 & 52.9 & 46.3 & 47.5 & 58.0 & 56.2 \\
		\checkmark & \checkmark & - & - & 51.1 & 48.9 & 61.7 & 57.5 & 46.8 & 48.1 & 58.1 & 56.0 \\
		\checkmark & - & \checkmark & - & 51.8 & 49.5 & 62.7 & 58.3 & 47.2 & 48.7 & 58.7 & 56.7 \\
		\checkmark & \checkmark & \checkmark & - & 52.7 & 50.2 & 64.2 & 59.6 & 48.0 & 50.5 & 60.8 & 58.6 \\
		\checkmark & \checkmark & \checkmark & \checkmark & 52.6 & 50.2 & 63.9 & 59.5 & 48.0 & 50.5 & 60.6 & 58.3 \\\hline
	\end{tabular}
\end{table*}

\section{Conclusion}
\label{sec:con}
In this work, we propose a simple and efficient framework for zero-shot sketch-based image retrieval (ZS-SBIR). The model is trained in an end-to-end fashion with three introduced losses: \textit{domain-aware quadruplet loss}, \textit{semantic classification loss} and \textit{semantic knowledge preservation loss}. The domain-aware quadruplet loss addresses the issue of domain-imbalance that occurrs using the vanilla triplet loss that is frequently used to reduce the domain gap and learn a shared low-dimension feature space. In addition, categorical semantic classification is also used to learn semantic features. To enhance the zero-shot ability of the learned model, the semantic knowledge preservation loss is introduced. This loss is formulated to prevent the rich knowledge learned from the ImageNet dataset from being forgotten during fine-tuning of the pre-trained ImageNet model that is used by the network. Experiments on three challenging ZS-SBIR datasets show that the proposed framework is more efficient and effective than related works. Moreover, extensive ablation studies show each introduced loss brings non-trivial improvements and contributes to the state-of-the-art performance.

\bibliographystyle{elsarticle-num}
\bibliography{egbib}

\end{document}